\pdfoutput=1

\documentclass[11pt]{article}

\usepackage{acl}

\usepackage{times}
\usepackage{latexsym}

\usepackage[T1]{fontenc}

\usepackage[utf8]{inputenc}

\usepackage{microtype}

\usepackage{inconsolata}

\usepackage{tcolorbox} 
\usepackage{xcolor}    
\usepackage{hyperref}  
\usepackage{cite}      
\usepackage{adjustbox}
\usepackage{algorithm}
\usepackage{algorithmic}
\usepackage{amsmath}
\usepackage{array}
\usepackage{booktabs}
\usepackage{colortbl}
\usepackage{cleveref}
\usepackage{enumitem}
\usepackage{float}
\usepackage{graphicx}

\newcommand{\methodfullnameind}{Penalty-Adjusted Type-Token Ratio}
\newcommand{\methodind}{PATTR}

\newcommand{\methodpair}{P-CAD}

\newcolumntype{L}[1]{>{\raggedright\let\newline\\\arraybackslash\hspace{0pt}}m{#1}}
\newcolumntype{C}[1]{>{\centering\let\newline\\\arraybackslash\hspace{0pt}}m{#1}}
\newcolumntype{R}[1]{>{\raggedleft\let\newline\\\arraybackslash\hspace{0pt}}m{#1}}

\newcommand{\parens}[1]{\left(#1\right)}
\newcommand{\braces}[1]{\left\{#1\right\}}
\newcommand{\bracks}[1]{\left[#1\right]}
\newcommand{\modulus}[1]{\left\vert#1\right\vert}
\newcommand{\norm}[1]{\left\Vert#1\right\Vert}

\title{A Penalty Goes a Long Way: Measuring Lexical Diversity in Synthetic Texts Under Prompt-Influenced Length Variations}

\author{
  Vijeta Deshpande$^{1}$ \quad Ishita Dasgupta$^{2}$ \quad Uttaran Bhattacharya$^{2}$ \\
  \textbf{Somdeb Sarkhel}$^{2}$ \quad \textbf{Saayan Mitra}$^{2}$ \quad \textbf{Anna Rumshisky}$^{1}$ \\
  \\
  $^{1}$University of Massachusetts Lowell \quad
  $^{2}$Adobe Inc. \\
  \texttt{vijeta\_deshpande@student.uml.edu}
}

\begin{document}
\maketitle

\begin{abstract}
Synthetic text generated by Large Language Models (LLMs) is increasingly used for further training and improvement of LLMs. Diversity is crucial for the effectiveness of synthetic data, and researchers rely on prompt engineering to improve diversity. However, the impact of prompt variations on response text length, and, more importantly, the consequential effect on lexical diversity measurements, remain underexplored. In this work, we propose \methodfullnameind\ (\methodind), a diversity metric robust to length variations.
We generate a large synthetic corpus of over 20M words using seven models from the LLaMA, OLMo, and Phi families, focusing on a creative writing task of video script generation, where diversity is crucial. We evaluate per-response lexical diversity using \methodind\ and compare it against existing metrics of Moving-Average TTR (MATTR) and Compression Ratio (CR).
Our analysis highlights how text length variations introduce biases favoring shorter responses. Unlike existing metrics, \methodind\ explicitly considers the task-specific target response length ($L_T$) to effectively mitigate length biases. We further demonstrate the utility of \methodind\ in filtering the top-10/100/1,000 most lexically diverse responses, showing that it consistently outperforms MATTR and CR by 
yielding on par or better diversity with high adherence to $L_T$.

\end{abstract}


\section{Introduction}
\label{sec:intro}

The rapid scaling of transformer-based language models has led to significant improvements in the quality of synthetically generated texts, often making them optically indistinguishable from human-written content \citep{orwig2024language, wu2025survey}.
Consequently, researchers are now leveraging synthetic text corpora for further training and refinement of large language models (LLMs) \citep{wang2022self, xu2024magpie, li2024synthetic, liu2024best, abdin2024phi, long2024llms, huggingface2024cosmopedia}.
However, diversity remains a crucial factor in determining the effectiveness of such synthetic data in model learning \citep{bukharin2023data, yu2024diversify}. Repeated training on synthetic data can reduce diversity, ultimately leading to model collapse \citep{guo2023curious, seddik2024bad, herel2024collapse}. Compromised diversity in synthetic texts \citep{shaib2024detection} also reduces collective diversity in human-LLM collaborative writing \citep{padmakumar2023does, doshi2024generative}.
To counteract this and enhance diversity in synthetic texts, researchers have extensively employed prompt engineering techniques \citep{long2024llms, ge2024scaling, huggingface2024cosmopedia}.
Yet, a critical but underexplored aspect remains: \textit{how does prompt engineering impact length variations in generated response, and how does length variation influence diversity measurement?}
While diversity in textual content encompasses multiple dimensions, including lexical, syntactical, and semantic, we focus on lexical diversity in this work, given its easier computational tractability.
The dependency of lexical diversity metrics on text length has been a long-standing challenge \citep{metric_mattr_ref, mccarthy2010mtld, metric_comp_ratio_ref}. Following Herdan-Heap's law, unique words in a corpus grow slower than total words, resulting in a higher proportion of unique words in shorter texts. Consequently, diversity metrics such as Type-Token Ratio (TTR) and Compression Ratio (CR) are inherently biased toward shorter texts \citep{mccarthy2010mtld} (see \Cref{sec:length_bias}). In this work, we observe that prompt variations can significantly impact response text length (see \Cref{tab:app:len_percentile}). Therefore, appropriate measurement of diversity in longer texts is particularly important, raising the need for a text-length-agnostic approach to measuring the diversity of synthetic texts.
To this end, we introduce a penalty term to modify TTR values, making them more robust to text length variations. We compute the penalty as the absolute difference between the target and the actual text lengths, which we then incorporate into the denominator of the TTR formulation. We refer to this modified metric as \methodfullnameind\ (\methodind). \methodind\ also explicitly considers task-specific target lengths ($L_T$), for example, $L_T=1{,}000$ words for essay writing or $L_T=200$ words for short story generation, to mitigate length biases. This flexibility allows \methodind\ to account for length variations while maintaining meaningful diversity measurements.


To evaluate the effectiveness of \methodind, we generate a large synthetic corpus of over 20M words using seven language models (LMs) from the LLaMA, OLMo, and Phi families. We focus on a creative writing task of video script generation, which encourages abstractive text generation and naturally leads to wide variations in response text lengths, thereby making it a suitable testbed for evaluating diversity metrics.
Each LM generates 12,000 video scripts by systematically varying three key components of the input prompt: instructions (10 unique values), style (10 unique values), and user prompt (120 unique values). First, we demonstrate how $L_T$ can be leveraged to smoothly control the bias towards shorter responses. We then assess the effectiveness of \methodind\ in filtering diverse responses. Across top-10/100/1,000 selections, \methodind\ outperforms MATTR and CR, yielding on par or better diversity (measured with ROUGE, BLEU, entropy, n-gram diversity, and Wasserstein distance) 
, for the filtered examples.

In summary, our main contributions include:  
\begin{itemize}
    \item \methodfullnameind\ (\methodind), consisting of a penalty on the response text length that effectively mitigates length bias, remains robust to response text length variations, and enhances the filtering of synthetic corpora for maximizing diversity.
    \item A dataset along with diversity measurements to facilitate further research on the impact of prompting on response text lengths and its influence on diversity metrics.
\end{itemize}

\section{Related Work}
\label{sec:related_work}

\paragraph{Measuring Text Diversity at Scale.}
Measuring text diversity is a well-studied topic \citep{metric_ttr_ref, metric_root_ttr_ref, metric_maas_ttr_ref, metric_mattr_ref, metric_hd_d_ref, metric_mtld_ref, metric_bert_score, metric_comp_ratio_ref, padmakumar2023does, salkar2022self}. Multiple variations on the idea of Type-Token-Ratio (TTR) have been proposed to measure diversity in a text string \citep{metric_root_ttr_ref, metric_mattr_ref, metric_maas_ttr_ref}. On the other hand, the idea of pairwise comparison has been explored to measure similarity (or the inverse diversity) within a collection of text strings \citep{metric_bert_score, padmakumar2023does}. One limitation of pairwise comparison methods is the quadratic increase in the runtime, leading to limited applicability in evaluating the diversity of large corpora \citep{metric_comp_ratio_ref}. In a recent study, \citet{metric_comp_ratio_ref} highlight the suitability of compression ratio for measuring diversity at scale. The runtimes of TTR-based metrics grow linearly with the length of the string and the number of strings. However, certain versions such as MTLD \citep{metric_mtld_ref} need multiple passes and can incur additional costs. 

\paragraph{Effect of Text Length.}
Following Herdan-Heap's law, it is well-known that vocabulary size grows sublinearly with increasing text lengths \citep{herdan1960type}. This phenomenon presents a challenge for lexical diversity metrics, often introducing a bias of better diversity towards shorter texts \citep{covington2010cutting, mccarthy2010mtld}. \citet{metric_comp_ratio_ref} highlighted a strong positive correlation between pairwise similarity scores and text lengths. To mitigate this length dependency, prior studies have explored techniques such as frequency correction, logarithmic transformations, text truncation, and moving averages \citep{mccarthy2010mtld, metric_mattr_ref, metric_comp_ratio_ref}. In this work, we introduce and investigate the utility of a penalty term in TTR that varies diversity scores non-linearly with changes in the text length.

\begin{figure*}[t]
    \centering
    \includegraphics[width=\textwidth]{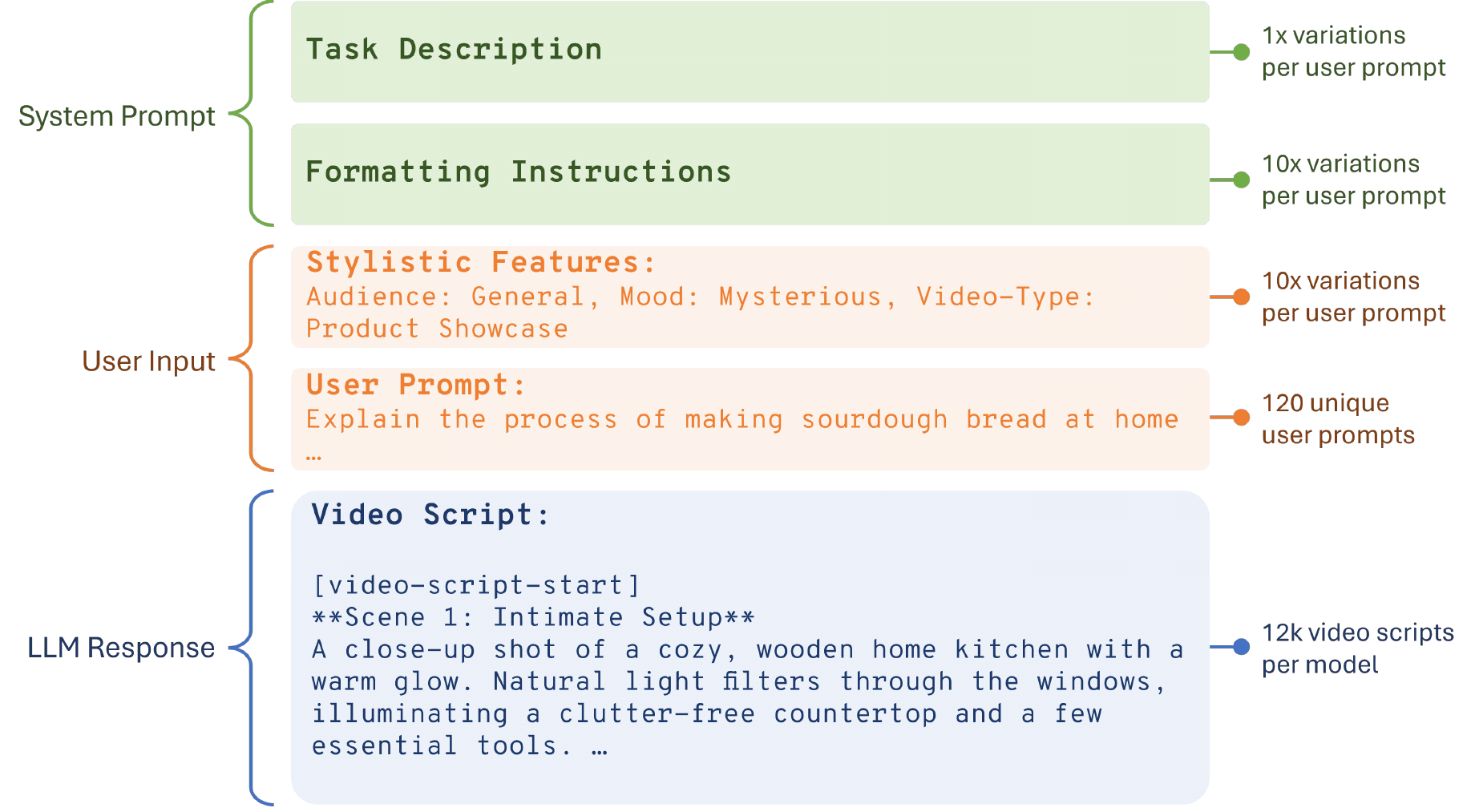}
    \caption{\textbf{Task example and synthetic data generation.} We show an example of the video script generation task and highlight key aspects of the synthetic dataset we generate based on this task.}
    \label{fig:task_example}
\end{figure*}

\paragraph{Impact of LLMs on Text Diversity.}
With the rapid popularization of LLM-powered chatbots for various writing tasks, maintaining a high quality of synthetically generated texts is of significant importance. Prior studies have highlighted a lack of diversity in synthetic texts \citep{padmakumar2023does, kirk2023understanding, shaib2024detection}, which, in turn, affects human writing when users collaborate with LLMs \citep{padmakumar2023does, doshi2024generative}. Additionally, the growing reliance on synthetic text for training LLMs has been shown to negatively impact model development, exacerbating the loss of diversity \citep{guo2023curious, seddik2024bad, herel2024collapse}. To counteract these effects, significant prompt engineering efforts are employed during synthetic data curation to maintain a desired level of diversity \citep{long2024llms, huggingface2024cosmopedia}. However, modifying prompts to enhance diversity can also lead to substantial variations in response text lengths, which reverts to the issue of diversity measurements depending on text length. In this work, we propose a diversity metric that explicitly accounts for such prompt-induced variations across a wide range of synthetic text lengths.


\section{Approach}
\label{sec:approach}
In this section, we elaborate on the synthetic data generation procedure, our proposed diversity metric, and how we evaluate diversity metrics.


\subsection{Synthetic Data Generation}

To best demonstrate the effects of variations due to prompting, we require a task that involves a high degree of diversity in model-generated responses. To this end, we focus on a creative writing task of generating video scripts based on user requests and task-specific instructions. To systematically study the impact of input variations, we decompose the model input, comprising of instructions and user requests, into four components. This structured approach enables us to introduce controlled perturbations and assess their effects on key properties of the generated scripts, such as length and diversity. The following paragraphs provide a detailed breakdown of the input structure and its components.

\paragraph{Model Input.}
We structure the model input into four components in a fixed sequence: \textit{task description}, \textit{formatting instructions}, \textit{style}, and \textit{topic} (see \Cref{fig:task_example} for an overview). We assume the \textit{task description} and the \textit{formatting instructions} are predefined by the NLP practitioner, and the \textit{style} and \textit{topic} are specified by the user.
The \textit{task description} provides a concise overview of the video script generation task and remains fixed across all experiments. The \textit{formatting instructions} outline specific guidelines for structuring the generated script, such as writing in a scene-by-scene format or summarizing the user request before generating the script. We define ten distinct formatting instructions and introduce them incrementally (including one with zero instructions), using variations, such as \texttt{<first-1>}, \texttt{<first-3>}, \texttt{<first-9>}, and so on.
The \textit{style} input tailors the video script to specific audience, mood, and video types. For each of these categories, we curate a set of five predefined values and sample one per category to simulate diverse user requests. In total, we generate 10 style variations for each user prompt for our analysis. The \textit{topic} input represents the user-provided request (\textit{i.e.}, user prompt) for generating a video script. We consider a diverse set of 120 topics, including both human-written and synthetic prompts. We carefully curate this set to ensure broad subject coverage and variations in prompt length, ranging from single words to 2-3 sentences.
We provide all the exact variations in Appendix \ref{sec:app:prompts}.

\paragraph{Special Tokens.}
We structure the combined input using a chat template, incorporating role-specific special tokens. We assign the task description and formatting instructions to the system role and the style and topic inputs to the user role. We append special tokens using the default tokenizer-specific chat templates available in the Hugging Face library\footnote{\url{https://huggingface.co/}}. For consistency, we refer to the final template-wrapped string as the model input, and we specifically refer to the topic subpart of the model input as the user prompt. Therefore, we can write the model input as
\begin{equation}
    \mathbf{x} = (x_0, x_1, ..., x_{L-1}),
\end{equation}
where each $x_i \in \mathbf{x}$ is a token such that $0 \leq x_i \leq |V| \ \forall i \in \{0, L-1 \}$, $V$ being the set of all tokens in the model vocabulary, and $L$ being the input sequence length.

\paragraph{Model Output.}
The generated script at the output is a sequence of tokens sampled from a language model $\pi(\cdot;\theta)$ conditioned on the input $\mathbf{x}$. We can write it as 
\begin{equation}
    \mathbf{y} = (y_0, y_1, ..., y_{M-1}),
\end{equation}
such that
\begin{equation}
    y_k \sim \pi(\mathbf{x}, y_{0:k-1}; \theta),
\end{equation}
where each $y_k \in \mathbf{y}$ is a token such that $0 \leq y_k \leq |V| \ \forall k \in \{0, M-1 \}$, $M$ being the output sequence length. We represent the language model using $\pi(\cdot; \theta)$, $\theta$ being the trainable parameters.

\paragraph{Models and Inference.}
With a fixed task description, 10 variations in formatting instructions and style inputs, and 120 unique user prompts (the topic component of the model input), we generate a total of 12,000 unique model inputs. We use this set of prompts to generate video-scripts from 11 language models, including OLMo-2 (7B and 13B), Llama-3.1 (8B), Llama-3.2 (1B and 3B), and Phi-3 (Mini and Medium).
Our analysis focuses exclusively on the instruct-tuned versions of these models. This diverse selection allows us to explore different categories of LMs: models trained with extensive synthetic data (Phi)
and standard dense autoregressive models (OLMo and Llama).
For each model, we generate 12,000 video scripts using temperature-based sampling with a temperature value of 1.0 and a fixed seed of zero. We perform inference in batches on NVIDIA RTX A6000 and 3090 GPUs, depending on availability. To ensure consistency, we clean the generated scripts by identifying sequences that include the assistant role's end token. The final dataset comprises over 100,000 cleaned scripts, totaling more than 50 million words (measured with whitespace-separated words).

\begin{figure*}[t]
    \centering
    \includegraphics[width=0.9\textwidth]{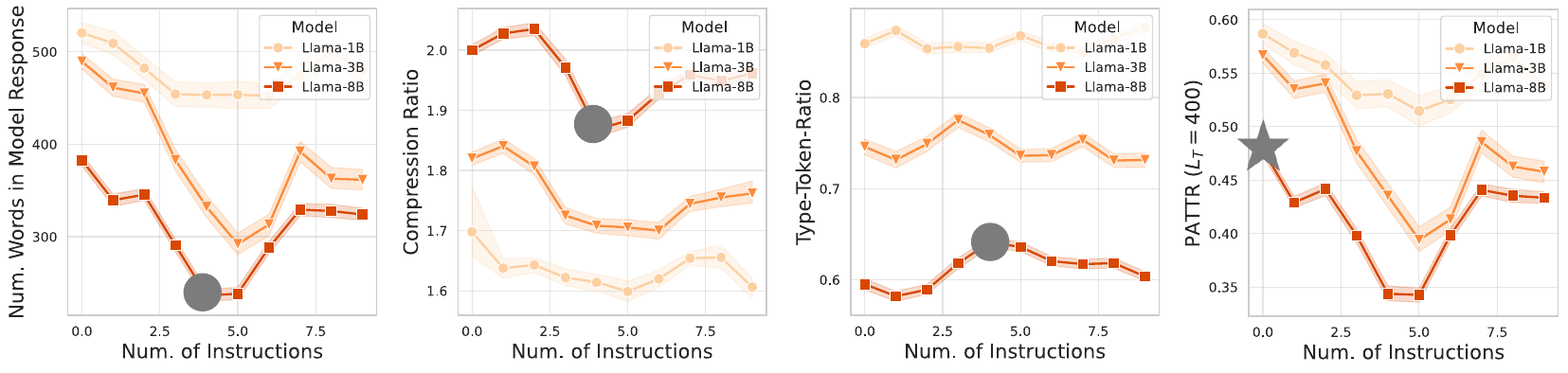}
    \caption{\textbf{Variations in response length and diversity scores w.r.t. the number of instructions to LMs.} \textit{Left to right $y$-axis:} response length variations, diversity scores using Compression Ratio (CR), using Type-Token Ratio (TTR), and using \methodind. CR (lower $\Rightarrow$ more diverse) and TTR (higher $\Rightarrow$ more diverse) favor the shortest responses (gray dot, corresponding to $\sim4$ instructions in this experiment) but \methodind\ (higher $\Rightarrow$ more diverse) considers the target length ($L_T=400$ in this experiment) and penalizes responses accordingly. Notably, \methodind\ identifies natural responses (gray star, zero instructions) as the most diverse.}
    \label{fig:variation_in_response_len}
\end{figure*}


\subsection{Proposed Diversity Metric}
\label{sec:metric}

Common diversity metrics such as Type-Token Ratio (TTR), Compression Ratio (CR), and those based on the Jaccard index exhibit a length bias, often identifying shorter texts as more diverse (see \Cref{tab:win_rate_small}). This bias poses a significant challenge, particularly in applications that require the generation of texts of specific lengths, such as writing a 1,000-word essay. In such cases, TTR and CR could incorrectly rank shorter essays as more diverse, leading to misleading evaluations. To address the limitations of these traditional metrics and ensure that the diversity measurements account for task-specific length constraints, we propose adjustments to TTR 
that mitigate length biases while preserving the integrity of diversity assessment.

\paragraph{Notation.}
For any text string, we denote the sequence of whitespace-separated words as a list $w = \bracks{w_0, \dots, w_{L-1}}$, where $L$ is the total number of words. We represent the number of unique words in $w$ as $\text{set}\parens{w} = \braces{u \mid u \in w}$. We can then define the Type-Token Ratio (TTR) scores as
\begin{equation}
    TTR\parens{w} = \frac{\modulus{\text{set}\parens{w}}}{\modulus{w}},
\end{equation}
where $\modulus{\ \cdot \ }$ denotes the size of the set or the list.

\paragraph{Penalty-Adjusted Metric.} 
To address the length bias in TTR,
we introduce a penalty term to adjust the scores based on the deviation from the target length for the given task, as

\begin{equation}
    P\parens{L, L_T} = \norm{L - L_T}_1,
\end{equation}
where $L$ denotes the number of whitespace-separated words in the generated text, and $L_T$ denotes the target length for the task. The $L_T$ value is a user-specific parameter that can be set according to task requirements (\textit{e.g.}, 1,000 for an essay or 200 for a short story). We incorporate this penalty term into the TTR formula to define the Penalty-Adjusted TTR (\methodind) as
\begin{equation}
    \methodind\parens{w, L_T} = \frac{\modulus{\text{set}\parens{w}}}{\modulus{w} + P\parens{\modulus{w}, L_T}}.
    \label{eqn:pattr}
\end{equation}

The penalty values increase linearly with the deviation from $L_T$, causing the denominator in 
\Cref{eqn:pattr} 
to increase accordingly. This, in turn, reduces the final diversity score. The absolute difference ensures that the penalty is applied to both shorter and longer texts relative to $L_T$. This bidirectional penalty, along with the flexibility to adjust $L_T$, are salient features of \methodind.

\subsection{Evaluation}
\label{sec:evaluation}

The primary objective of this work is to enhance the measurement of text diversity. Accordingly, our evaluation consists of two key components: assessing text diversity and evaluating the effectiveness of diversity metrics. For the first component, we employ
our proposed metric \methodind\ as well as conventional diversity measures, including TTR, MATTR, and CR. 
Originally introduced to measure corpus-level diversity \citep{metric_comp_ratio_ref}, the CR metric is repurposed in our study to assess sample-level diversity by treating each response as a single-document corpus. 
For the second component, we analyze the length bias of diversity metrics and assess their suitability for filtering text corpus to select the most diverse samples.

\begin{table}[t]
\centering
\begin{tabular}{@{}lc@{}}
\toprule
\textbf{$\boldsymbol{L_T}$} & \textbf{Correlation Coeff. (p-value)} $^{***}$ \\
\midrule
100  & $-0.4197$ \\
275  & $+0.0329$ \\
400  & $+0.9104$ \\
\bottomrule
\end{tabular}
\caption{\textbf{Correlation between PATTR and response length.} Spearman correlation between PATTR and the response length (in whitespace-separated words) varies with the target response length ($L_T$), exhibiting negative, neutral, and positive trends as $L_T$ increases. $^{***}$ denotes significance at $p < 0.001$.}
\label{tab:len-correlation}
\end{table}


\begin{table*}[t]
    \centering
    \begin{adjustbox}{max width=\textwidth}
    \begin{tabular}{lC{0.5mm}cC{0.5mm}cC{0.5mm}ccc}
        \toprule
        && \textbf{CR} && \textbf{MATTR} && \multicolumn{3}{c}{\textbf{\methodind}} \\
        \textbf{Model} && $L=128$ && $W=32$ && \textbf{$L_T=200$} & \textbf{$L_T=400$} & \textbf{$L_T=600$} \\
        \cmidrule{1-1}\cmidrule{3-3}\cmidrule{5-5}\cmidrule{7-9}
        OLMo-2-13B      && $67.33$ && $38.33$ && $28.75$ & $0.58$ & $0.08$ \\
        Llama-3.1-8B    && $37.17$ && $20.17$ && $64.67$ & $0.58$ & $0.17$ \\
        Phi-3-med       && $43.50$ && $42.58$ && $90.00$ & $20.83$ & $1.92$ \\
        \bottomrule
    \end{tabular}
    \end{adjustbox}
    \caption{\textbf{Win rate for short responses.} We evaluate the tendency of diversity metrics to favor shorter responses. Given a pool of 10 model-generated responses for a fixed set of instructions and user prompts, we select the most diverse response using Compression Ratio (CR) (truncation length $L=128$ words), Moving Average Type-Token Ratio (MATTR) (window length $W=32$ words), and PATTR ($L_T \in \{200, 400, 600 \}$). The win rate represents the percentage of selected responses with a word count below the $25^{th}$ percentile of the pool. Higher win rates indicate a stronger bias toward shorter sequences. PATTR, relying on $L_T$, can achieve better robustness to length bias.} 
    \label{tab:win_rate_small}
\end{table*}

\paragraph{Length Bias.}
To quantify length bias, we measure the win rate of short sequences. Specifically, we compute the win rate by analyzing video scripts generated for the same instruction and user prompt (10 scripts per prompt with varying style inputs). We rank these 10 scripts using \methodind, MATTR, or CR (one at a time) and check whether the script with the highest diversity score falls within the first quartile of script length, \textit{i.e.}, at or below the $25^{th}$ percentile of video-script length. We measure length as the number of whitespace-separated words, and determine the $25^{th}$ percentile within the pool of 10 scripts being compared. We record a win if the top-ranked script is within this first quartile. We then compute the average win rate across 1,200 samples per model, derived from 10 versions of instructions and 120 user prompts.

\paragraph{Corpus Diversity.}
We also evaluate the effectiveness of diversity metrics in filtering a corpus to optimize diversity.
For filtering, we sort the model responses based on their \methodind, MATTR, and CR scores, and select the corresponding top 10, 100, or 1,000 most diverse responses. We assess the overall similarity within the filtered corpus using ROUGE-1/2/L \citep{lin2004rouge} and BLEU \citep{papineni2002bleu}, which have been commonly used in previous studies as indicators of diversity \citep{padmakumar2023does, metric_comp_ratio_ref}, and are also known as homogenization scores. We use the implementations of ROUGE-1/2/L and BLEU scores in the Hugging Face library with default parameters. An ideal diversity metric should produce a filtered corpus with lower homogenization scores.
Since ROUGE-1/2/L, and BLEU are all pairwise comparison metrics, 
the runtime becomes quadratic in the size of the corpus. Thus, to keep the evaluation time manageable, we calculate similarity values for up to $1{,}000$ randomly sampled pairs (\textit{e.g.}, for top-100 selection, we select $1{,}000$ out of $100(100-1)/2=4{,}950$ unique pairs).
We also report the average per-token entropy of the filtered corpus using the SmolLM2-135M/360M/1.7B causal language models, with higher entropy values indicating greater diversity. To estimate the corpus-level diversity, we compute the $N$-gram diversity, $\sum_{n=1}^{N} \left( \text{unique $n$-grams / total $n$-grams} \right)$ \citep{metric_comp_ratio_ref}, and the Wasserstein distance, the sum of absolute differences between two CDFs \citep{vaserstein1969markov}.



\section{Results}
\label{sec:results}

In this analysis, we focus on effective diversity measurement for the synthetic text, \textit{i.e.}, text sampled from LMs. We particularly note the variations in the model response length and its effect on diversity measurements with conventional metrics. We also present the capability of \methodind\ to overcome these challenges, and show the effectiveness of \methodind\ in identifying highly diverse texts in a large corpus. 

While our analysis is more focused on synthetic text, it is important to note that our approach is equally applicable for diversity measurements in any form of text, such as fully human-written text and human-LLM collaboratively-written text.

\subsection{Length Correlation of \methodind}
Prior work has shown that diversity metrics such as MATTR and CR correlate strongly with text length and estimate higher diversity for shorter texts \citep{metric_comp_ratio_ref}. For \methodind, we show that its correlation with text length \textit{varies} based on the
target response length $L_T$. Using responses generated by the Llama-3.1-8B model as an example, 
we observe that setting $L_T$ to 100, 275, and 400 respectively yields a strong negative, near-zero, and a strong positive correlation with the response length (Table \ref{tab:len-correlation}). 
We illustrate this positive correlation in \Cref{fig:variation_in_response_len}, where PATTR values (right-most figure) have a trend similar to that of response length (left-most figure). We report the corresponding length distributions of the model responses in Table~\ref{tab:app:len_percentile}.

\begin{figure*}[t]
    \centering
    \includegraphics[width=\textwidth]{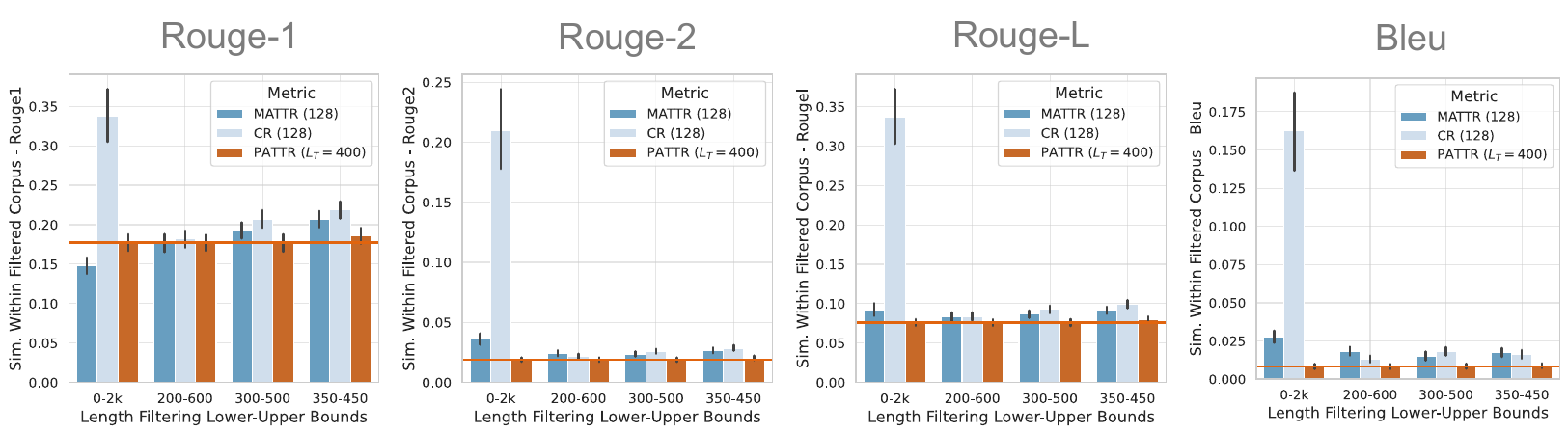}
    \caption{\textbf{Evaluation of top-10 diverse examples with pairwise similarity scores.} Average pairwise similarity scores (ROUGE-1/2/L, BLEU) for the top-10 diverse examples selected by \methodind\ ($L_T=400$), MATTR (window length of 128 words), and CR (truncation length of 128 words). The $x$-axis represents different length constraints (\textit{e.g.}, 200-600: 200 $\leq$ word count $\leq$ 600). The $y$-axis shows similarity scores (lower values indicate greater diversity). We average the similarity scores for all seven models. The horizontal orange line represents \methodind\ without length filtering. Except for ROUGE-1 with 0-2K filter, \methodind\ consistently outperforms MATTR and CR.}
    \label{fig:pattr_data_filter_top_10}
\end{figure*}

\subsection{A Solution for Length Bias}
\label{sec:length_bias}

\paragraph{Why is Length Bias a Challenge?}
The first step in generating synthetic text is prompting. LMs, especially the instruction-tuned versions, are developed to understand and address all the tasks mentioned in the prompt. Hence, prompt variations are expected to change the model response and, consequently, affect the response length. However, the extent to which prompt modifications influence response length remains hard to track in NLP research. Therefore, we investigate changes in the response length with our structured prompting setup. In \Cref{fig:variation_in_response_len} (leftmost sub-figure), we report the variations in response length for LLaMA models (8B, 3B, and 1B) and find that increasing the number of instructions leads to wide variations in response lengths (\textit{e.g.}, $\sim$200-400 words for LLaMA-8B). Within this range, conventional diversity metrics such as Type-Token Ratio (TTR) and Compression Ratio (CR) exhibit a length bias in favor of shorter responses. Notably, in \Cref{fig:variation_in_response_len}, the shortest response (marked with a gray dot) achieves the highest TTR and lowest CR values. We observe this trend consistently and prominently across all investigated models except for Phi-3-medium (see \Cref{fig:app:resp_len_vs_num_inst}). These findings highlight how response length variations, coupled with biased diversity metrics, make identifying diverse data extremely challenging. Further, having such variations in the response length makes our setup an ideal test bed for evaluating diversity measurements.

\paragraph{Can Length Penalty Overcome Length Bias?}
To mitigate the impact of response length variations on diversity measurement, we introduced a length penalty term to TTR to compute \methodind\ (see \Cref{sec:metric}). As we show in \Cref{fig:variation_in_response_len}, unlike TTR and CR, the highest \methodind\ score (marked with a gray star) does not correspond to the shortest response, demonstrating its reduced sensitivity to length bias. The length penalty term in \methodind\ allows practitioners to adapt diversity measurements to task-specific length requirements. \Cref{tab:win_rate_small} illustrates how \methodind\ leverages the target length ($L_T$) to control length bias. We compare \methodind\ against truncated CR (truncation length $L=$ first 128 words) and MATTR (window size $W=$ 32 words), and show that larger values of $L_T$ lead to \methodind's reduced win rates for shorter sequences.

\subsection{Application of \methodind\ in Data Filtering}
Having established the key features of \methodind, we apply it to filter a synthetic text corpus. Filtering refers to ranking texts based on diversity scores and selecting the top-$k$ samples. We use \methodind, MATTR, and CR to rank 12,000 video scripts generated by each model. From these rankings, we create filtered datasets by selecting the top-10, top-100, and top-1,000 samples. To assess the diversity of the filtered sets, we compute homogenization scores using ROUGE-1/2/L and BLEU for the top-10/100/1,000 selections, and use entropy to evaluate top-1,000 selections only. We generate the top-$k$ selections for all seven models in our experiments and report the homogenization scores and entropy values averaged across models and top-$k$ examples.  

Since \methodind\ incorporates task-specific target length information, it may have an inherent advantage over MATTR and CR. To ensure a fair comparison, we augment MATTR and CR-based filtering with length-based constraints. Specifically, when the target length $L_T=400$ words, we first exclude video scripts that fall outside predefined length (\#~words) ranges, 0-2K, 200-600, 300-500, and 350-450, before applying MATTR or CR-based ranking. In the absence of \methodind, such length-based filtering strategies provide a practical alternative for mitigating the inherent bias of MATTR and CR toward shorter responses.

\paragraph{Can \methodind\ Find More Diverse Samples?} 

\begin{figure*}[t]
    \centering
    \includegraphics[width=\textwidth]{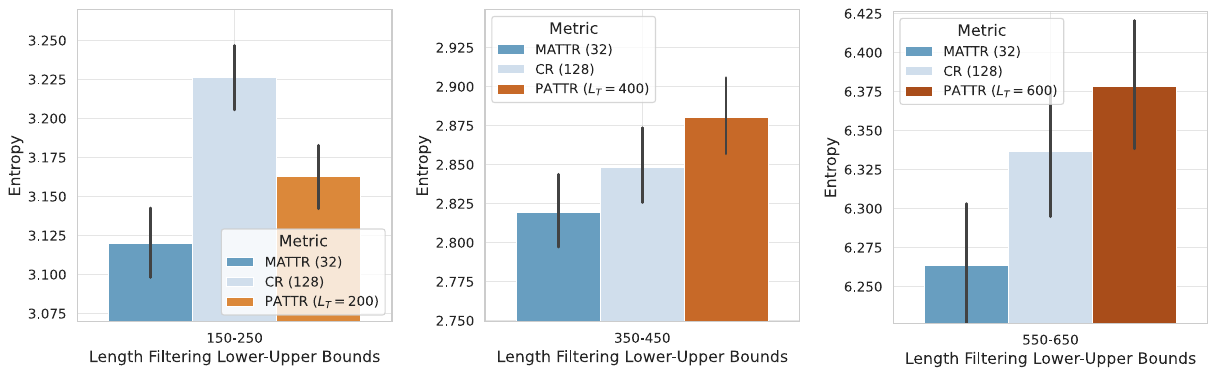}
    \caption{\textbf{Evaluation of top-1,000 diverse examples with entropy.} We measure the diversity of top-1,000 examples selected by \methodind\ ($L_T \in \{200, 400, 600\}$), MATTR (32-word window), and CR (first 128 words) with entropy (based on SmolLM2-1.7B). Higher values of entropy represent a more diverse set of video scripts.}
    \label{fig:pattr_data_filter_top_1000_entropy}
\end{figure*}

\paragraph{Evaluation-1: Pairwise Similarity.}
\Cref{fig:pattr_data_filter_top_10} presents the evaluation of the top-10 selections by \methodind\ ($L_T=400$), MATTR, and CR. Without length constraints (represented by the 0-2K point on the $x$-axis and the horizontal orange line), \methodind\ consistently outperforms MATTR and CR across all scenarios except ROUGE-1 with no length constraints. This highlights the robust nature of \methodind\ against variations in the response length. 
Overall, we evaluate 16 scenarios, including four length constraints, each with four similarity metrics. As we report in \Cref{fig:app:pattr_data_filter_top_100}, \methodind\ outperforms both CR and MATTR in 14, 12, and 15 of the 16 scenarios, respectively, for top-10, top-100, and top-1,000 selections.\footnote{Note that, with the increase $k$ of top-k selection along with length filtering, the corpus selected based on different metrics loses mutual exclusivity. Hence, the corpus-level similarity metric approach to the same value for all metrics (used for filtering) with increasing $k$ and/or narrowing range of the length filtering.} 

\paragraph{Evaluation-2: Entropy.} 
We extend our evaluation by measuring the average entropy for the top-1,000 video scripts selected by \methodind, MATTR and CR. Specifically, we employ SmolLM2 models \citep{allal2025smollm2}, which have a pretraining context length of 2,048 tokens.\footnote{While the authors extend the context length to 8K tokens for the 1.7B model, it is unclear if similar extensions have been applied to the smaller 135M and 350M checkpoints.}
Since all generated scripts fall within the 2,048-token limit, we compute the average per-token entropy for the entire video script without the moving window approach.
Higher entropy values indicate greater diversity, but the text length significantly affects entropy measurements. Within the context length limit, entropy naturally decreases as text length increases. Thus, selecting a higher proportion of shorter scripts can inflate entropy values, leading to a misleading impression of diversity. To mitigate this effect, we apply stricter length constraints: 150-250, 350-450, and 550-650 words. For each constraint, we compare MATTR and CR against \methodind\ with $L_T=200$, $L_T=400$, and $L_T=600$, respectively. Importantly, we apply these constraints uniformly across all metrics to ensure a fair comparison.
\Cref{fig:pattr_data_filter_top_1000_entropy} presents entropy measurements using the SmolLM2-1.7B model, with additional results for the 135M and 360M checkpoints provided in Appendix \ref{sec:data_filtration}. For the 150-250 range, \methodind\ ($L_T=200$) achieves the second-highest entropy, slightly trailing CR. However, as the selection shifts toward longer video scripts (350-450 and 550-650 words), \methodind\ consistently results in a more diverse corpus. We also observe this trend with smaller SmolLM2 checkpoints (see \Cref{fig:app:pattr_data_filter_top_1000_entropy}), further validating the robustness of \methodind\ in selecting diverse responses across varying text lengths.

\paragraph{Evaluation-3: $\boldsymbol{N}$-Gram Diversity.} 
We further evaluate the diversity of top-10 responses using the $N$-gram diversity metric. Since $N$-gram diversity is a TTR-style measure applied to all $n$-grams up to $N$, it inherits the same length bias: shorter texts tend to have a higher proportion of unique $n$-grams and, consequently, inflated diversity scores. To control this, we apply a strict length filter and only consider responses between 350 and 450 words. For each model, we select its top-10 responses, compute 4- and 6-gram diversity, and report the average across seven models (see \Cref{tab:n_gram_diversity}). We also compute 4- and 6-gram diversity over POS-tag sequences to assess syntactic diversity. Interestingly, although \methodind's top-10 responses tend to be longer than those from MATTR or CR, they consistently show higher diversity than CR and are slightly below MATTR. Moreover, \methodind\ achieves the highest POS-based diversity despite its longer response length.

\begin{table}[t]
\centering
\small
\begin{tabular}{@{}lccc@{}}
\toprule
 & \textbf{CR} & \textbf{MATTR} & \textbf{PATTR} \\
\textbf{Metric} & $L=128$ & $W=128$ & $L_T=400$ \\
\midrule
Resp. Len       & $395.06$ & $392.84$ & $\mathbf{398.20}$ \\
\midrule
4-gram          & $3.53$   & $\mathbf{3.66}$   & $\underline{3.65}$   \\
6-gram          & $5.52$   & $\mathbf{5.65}$   & $\underline{5.64}$   \\
\midrule
4-gram (POS)    & $\mathbf{0.62}$   & $\underline{0.60}$   & $\mathbf{0.62}$   \\
6-gram (POS)    & $\underline{2.09}$   & $2.07$   & $\mathbf{2.12}$  \\
\bottomrule
\end{tabular}
\caption{\textbf{$\boldsymbol{N}$-gram diversity of top-10 selections under a length filter of 350-450 words.} We report 4- and 6-gram and POS-based diversity for top-10 examples selected by CR (128 words), MATTR (128-word window), and PATTR (400-word target). PATTR achieves comparable $N$-gram diversity and the highest POS diversity, despite selecting longer responses.}
\label{tab:n_gram_diversity}
\end{table}

\paragraph{Evaluation-4: Distance from the Most Diverse (Uniform) Distribution.} 
Lastly, we evaluate lexical diversity by comparing the cumulative vocabulary distributions of the top-10 responses selected by \methodind\ ($L_t = 400$), MATTR ($W = 128$), and CR ($L = 128$), without applying any length-based filtering. For each model and metric, we compute the empirical cumulative distribution over the vocabulary and compare it against a reference distribution induced by a uniform vocabulary usage.\footnote{A uniform distribution assumes that each word appears exactly once in the corpus, thereby achieving maximal lexical diversity.} We use the Wasserstein distance (also known as Earth Mover's Distance) \citep{vaserstein1969markov} to quantify the deviation from this ideal distribution. \methodind\ yields the lowest Wasserstein distance ($46.37$), indicating higher diversity compared to MATTR ($125.23$) and CR ($154.26$). These findings further demonstrate the effectiveness of \methodind\ in identifying lexically diverse responses.

\subsection{Sensitivity analysis for $\boldsymbol{L_T}$}
We conduct a sensitivity analysis on \methodind\ and MATTR using LLaMA-3.1 8B responses (see \Cref{tab:sensitivity_analysis_t_test_FULL} for OLMo-13B and Phi-med results), varying target length $L_T$ for \methodind\ and window length $W$ for MATTR. For each configuration, we compute \methodind\ and MATTR scores, rank the responses, and evaluate corpus-level diversity using the top-10 ranked outputs. We measure diversity via 45 inter-sample pairwise similarity scores from all unique response pairs from the top-10 selection, and perform independent t-tests to compare selections across metrics. A negative T-statistic indicates higher diversity for \methodind, while a positive difference in average length suggests that \methodind\ selects longer responses. Our findings in \Cref{tab:sensitivity_analysis_t_test} 
show that \methodind\ consistently selects longer responses and achieves higher corpus-level diversity based on ROUGE-L.
We also observe that varying the MATTR window size has minimal impact on the length of selected examples. Notably, \methodind\ selections are both longer and significantly more diverse ( Length Diff. $> 0$, T-statistic $< 0$, and $p < 0.001$). This provides strong evidence that \methodind\ is more effective at identifying diverse, length-aware outputs.

\begin{table}[t]
\centering
\small
\begin{tabular}{@{}cccc@{}}
\toprule
\textbf{PATTR} & \textbf{MATTR} & \textbf{Length}  & \textbf{ROUGE-L} \\
\textbf{$L_T$} & \textbf{$W$} & \textbf{Diff.} & T-stat. \\
\midrule
$100$  & $32$  & $93.6$   & $-5.95^{***}$ \\
$100$ & $128$ & $93.6$   & $-5.95^{***}$ \\
$100$  & $512$ & $93.6$   & $-5.73^{***}$ \\
\midrule
$275$  & $32$  & $275.0$  & $-9.24^{***}$ \\
$275$  & $128$ & $275.0$  & $-9.24^{***}$ \\
$275$  & $512$ & $275.0$  & $-9.16^{***}$ \\
\midrule
$400$  & $32$  & $411.7$  & $-10.96^{***}$ \\
$400$  & $128$ & $411.7$  & $-10.96^{***}$ \\
$400$  & $512$ & $411.7$  & $-10.95^{***}$ \\
\bottomrule
\end{tabular}
\caption{\textbf{Effect of variations in the target length.} Negative t-statistics indicate that top-10 responses selected by PATTR are more diverse compared to MATTR. A positive Length Diff. means these responses are also longer than MATTR's corresponding selections. $^{***}$ denotes significance at $p < 0.001$. The combination of (Length Diff. $> 0$), (T-stat $< 0$), and ($p < 0.001$) highlights cases where PATTR effectively selects more diverse and longer responses.}
\label{tab:sensitivity_analysis_t_test}
\end{table}

\section{Conclusion}
\label{sec:conclusion}


We introduced \methodfullnameind\ (\methodind), a penalty-adjusted extension of the Type-Token Ratio (TTR) designed to mitigate the inherent bias of conventional diversity metrics toward shorter responses. By incorporating the task-specific target length ($L_T$), \methodind\ provides a flexible mechanism for controlling length bias, addressing a key limitation observed in metrics such as TTR, MATTR, and CR. Through extensive experiments, we demonstrated that \methodind\ effectively enhances the filtering of synthetic corpora to maximize lexical diversity. Our results show that adjusting $L_T$ allows users to fine-tune diversity measurements based on task requirements, making \methodind\ a more adaptable and robust metric. Beyond our empirical findings, we contribute a large synthetic corpus annotated with diversity measurements to facilitate further research on the interplay between prompting, response length, and diversity metrics. We hope this resource will support future studies in improving diversity-aware evaluations for synthetic text generation.

\section*{Limitations}
\label{sec:limitations}
The proposed method requires a task-specific input: the target response length ($L_T$) in words. While this aligns well with structured creative writing tasks such as essay or short-story writing, where response length can be reasonably estimated or constrained, its applicability to more open-ended writing tasks may be limited. In such cases, practitioners can conduct sensitivity analyses with varying $L_T$ values to identify the most suitable setting for their task. Additionally, our work primarily focuses on lexical diversity, leaving the exploration of length penalties for syntactic and semantic diversity measurements as future research directions. Furthermore, we would like to note that similar to TTR and MATTR, \methodind\ also measures the diversity for one sample and does not consider inter-sample similarity or diversity when evaluating a corpus-level diversity. However, in our study, we find that \methodind-based corpus filtering results in better or comparable corpus diversity values (refer to Figures \ref{fig:pattr_data_filter_top_10}, \ref{fig:pattr_data_filter_top_1000_entropy}, and Tables \ref{tab:n_gram_diversity}, \ref{tab:sensitivity_analysis_t_test}). 
We also do not investigate the agreement of \methodind\ with human judgments. However, prior studies have shown low inter-annotator agreement in creative writing tasks \citep{gomez2023confederacy, chakrabarty2023creativity, chakrabarty2024can}, highlighting inherent preferential inconsistencies among human evaluators. Since this issue pertains to broader subjectivity in human assessments, it falls outside the scope of our study.

\section*{Ethical Considerations}
While our work analyzes the issues of measuring the lexical diversity of contents generated by language models, our proposed metric is not a surrogate for measuring the \textit{overall quality} of generated contents, and should be considered in combination with existing metrics of generation accuracy and fidelity, as applicable, when evaluating the performance of language models. Our proposed metric has also not been demonstrated to be a statistical indicator of other dimensions of diversity, such as syntactic and semantic, and, therefore, should be reported with the appropriate qualification. Furthermore, in writing this paper, we have used proprietary chatbots for text editing. No part of this paper is completely generated from any language model. 



\bibliography{custom}

\clearpage
\appendix

\counterwithin{equation}{section}
\counterwithin{figure}{section}
\counterwithin{table}{section}

\section{Prompts and Model Output}
\label{sec:app:prompts}
For our experiments, the model input comprises four distinct parts: task description, formatting instructions, stylistic features, and user prompt. The task description part of the model input briefly instructs the model of the task at hand, written as follows:

\begin{tcolorbox}[
    colback=gray!05,  
    colframe=black!50, 
    boxrule=0.5pt,     
    arc=2mm,           
    width=\linewidth,  
    leftrule=0.1mm,      
    rightrule=0.1mm,     
    bottomrule=0.1mm,    
    toprule=0.1mm,       
]

\texttt{\textbf{Task Description:}}

\texttt{You are a conversational assistant specializing in creating engaging and innovative video scripts for short videos (less than a minute long). Your task is to generate video scripts based on user-provided prompts and stylistic preferences.
You will receive a prompt from the user describing the main topic of the video, along with stylistic features that reflect the user's preferences. Your goal is to write a creative and engaging script for a short-video that aligns with both the user's topic and stylistic requirements.}

\end{tcolorbox}

The second part of the model inputs consists of the formatting instructions. Unlike the task description, this part is variable \textit{i.e.}, we vary the total number of formatting instructions included in the model input from zero to nine in our experiments. For $k$ total formatting instructions to be included in the model input, we select the first $k$ instruction from the following list (note, we do not sample the instruction).
\begin{tcolorbox}[
    colback=gray!05,  
    colframe=black!50, 
    boxrule=0.5pt,     
    arc=2mm,           
    width=\linewidth,  
    leftrule=0.1mm,      
    rightrule=0.1mm,     
    bottomrule=0.1mm,    
    toprule=0.1mm,       
]

\texttt{\textbf{List of Formatting Instructions:}}

\texttt{While generating the video script please strictly adhere to following formatting rules:}

\begin{enumerate}[label=\texttt{{\arabic*}.}]
    \item \texttt{Start the video script with [video-script-start] and after the last scene end with [video-script-end].}
    \item \texttt{The video script should be written in scene-by-scene format like [scene-1]: ..., [scene-2]: .... etc.}
    \item \texttt{Every scene must have a brief description of the scene. Do not exceed 30 words per scene.}
    \item \texttt{Generate five (5) or less scenes for the video script.}
    \item \texttt{For better readability, separate the scenes with a blank line.}
\end{enumerate} 

\end{tcolorbox}

\begin{tcolorbox}[
    colback=gray!05,  
    colframe=black!50, 
    boxrule=0.5pt,     
    arc=2mm,           
    width=\linewidth,  
    leftrule=0.1mm,      
    rightrule=0.1mm,     
    bottomrule=0.1mm,    
    toprule=0.1mm,       
]

\begin{enumerate}[label=\texttt{{\arabic*}.}, start=6]
    \item \texttt{Begin conversation with summarizing the user request in the prompt just to make sure you understand it correctly. This summary must appear before the [video-script-start] tag.}
    \item \texttt{You must conclude the video script with a call to action or a closing message such as asking to like, share and subscribe. This must appear after the last scene and before the [video-script-end] tag.}
    \item \texttt{If the user wants any specific changes to the script, ask them to provide feedback or suggestions. This must appear after the [video-script-end] tag.}
    \item \texttt{If the user likes the script, ask them to click on 'Create video' button.}
\end{enumerate}

\end{tcolorbox}

The third part of the model input consists of stylistic features. We define three categories of styles namely, audience, mood, and video type, and manually craft five possible values for each category. In each model input, we enter only one randomly sampled value per style category. In our experiments, we use 10 randomly sampled and unique stylistic features per user prompt. We list all style categories and their respective values below:
\begin{tcolorbox}[
    colback=gray!05,  
    colframe=black!50, 
    boxrule=0.5pt,     
    arc=2mm,           
    width=\linewidth,  
    leftrule=0.1mm,      
    rightrule=0.1mm,     
    bottomrule=0.1mm,    
    toprule=0.1mm,       
]
\texttt{\textbf{Style Categories and Values:}}

\begin{itemize}
    \item \texttt{Audience: Teenagers, Young Adults, Middle-aged Adults, Elderly, General Audience.}
    \item \texttt{Mood: Funny, Calm, Mysterious, Romantic, Motivational.}
    \item \texttt{Video Type: Reel, Time-lapse, Tutorial, Product Showcase, Interview}
\end{itemize}

\texttt{\textbf{An Example of Stylistic Features in Model Input:}}

\texttt{Stylistic Features: Audience: General Audience, Mood: Romantic, Video Type: Product Showcase}

\end{tcolorbox}

The last part of the model is the user prompt. We provide an example of the user prompt below:
\begin{tcolorbox}[
    colback=gray!05,  
    colframe=black!50, 
    boxrule=0.5pt,     
    arc=2mm,           
    width=\linewidth,  
    leftrule=0.1mm,      
    rightrule=0.1mm,     
    bottomrule=0.1mm,    
    toprule=0.1mm,       
]
\texttt{\textbf{An Example of User Prompt:}}

\texttt{Explain the process of making sourdough bread at home from scratch.}

\end{tcolorbox}

The final model input for the above user prompt with two instructions and the example stylistic features looks as follows:

\begin{tcolorbox}[
    colback=gray!05,  
    colframe=black!50, 
    boxrule=0.5pt,     
    arc=2mm,           
    width=\linewidth,  
    leftrule=0.1mm,      
    rightrule=0.1mm,     
    bottomrule=0.1mm,    
    toprule=0.1mm,       
]

\texttt{Task Description: You are a conversational assistant specializing in creating engaging and innovative video scripts for short videos (less than a minute long). Your task is to generate video scripts based on user-provided prompts and stylistic preferences. You will receive a prompt from the user describing the main topic of the video, along with stylistic features that reflect the user’s preferences. Your goal is to write a creative and engaging script for a short-video that aligns with both the user’s topic and stylistic requirements.}\\

\texttt{While generating the video script please strictly adhere to following formatting rules: 1. Start the video script with [video-script-start] and after the last scene end with [video-script-end]. 2. The video script should be written in scene-by-scene format like [scene-1]: ..., [scene-2]: .... etc.}\\

\texttt{Stylistic Features: Audience: General Audience, Mood: Romantic, Video Type: Product Showcase}\\

\texttt{User Prompt: Explain the process of making sourdough bread at home from scratch.}

\end{tcolorbox}

\begin{tcolorbox}[
    colback=gray!05,  
    colframe=black!50, 
    boxrule=0.5pt,     
    arc=2mm,           
    width=\linewidth,  
    leftrule=0.1mm,      
    rightrule=0.1mm,     
    bottomrule=0.1mm,    
    toprule=0.1mm,       
]

\texttt{[video-script-start]}\\

\texttt{[scene-1]}

\texttt{[intro music lightly plays]}

\texttt{Text on screen: ``Discover the Art of Sourdough''}

\texttt{Voiceover (calm and soothing): [Scene: A cozy kitchen, early morning sunlight filtering through the windows, gentle music playing in the background] ``Ever wondered how to craft a loaf that's as complex as the stories it can tell?''}

\texttt{[cut to TEXT ON SCREEN: ``Home-Baked Magic'']}\\

\texttt{[scene-2]}

\texttt{[Cut to Close-up of flour and water in a bowl]}

\texttt{Voiceover: ``Today, we dive into the ancient tradition of making sourdough bread.''}\\

\texttt{[scene-3]}

\texttt{[Text on screen: ``Step 1: Create the Starter'']}

\texttt{Voiceover [Scene: Flour in a jar]: ``Our journey begins with a simple mix of flour and water...''}

\texttt{[Cut to a hand pouring flour into a mason jar, slowly adding water]}

\texttt{Voiceover (guiding): ``This is our sourdough starter. Think of it as the heart of our bread.''}\\

\texttt{[scene-4]}

\texttt{[Text on screen: ``Day 1'']}

\texttt{Voiceover: ``Day one: Sealed and forgotten, our starter begins its magic...''}\\

\texttt{[scene-5]}

\texttt{[Time lapse montage. Jar with starter: morning, afternoon, evening -- gently bubbling, expanding]}

\texttt{Voiceover: ``Over the next few days, we watch and wait as natural yeast works its wonders...''}

\end{tcolorbox}

\begin{tcolorbox}[
    colback=gray!05,  
    colframe=black!50, 
    boxrule=0.5pt,     
    arc=2mm,           
    width=\linewidth,  
    leftrule=0.1mm,      
    rightrule=0.1mm,     
    bottomrule=0.1mm,    
    toprule=0.1mm,       
]

\texttt{[scene-6]}

\texttt{[Text on screen: ``Day 4'']}

\texttt{Voiceover: ``Revelation day! Our starter's ready to brew up our dough.''} \\

\texttt{[scene-7]}

\texttt{[Cut to kneaded dough gently pushed and folded]}

\texttt{Voiceover: ``Into our starter, we add more flour, salt, and the cycle of mixing, folding, and nurturing begins.''}\\

\texttt{[scene-8]}

\texttt{[Text on screen: ``Time for Rest'']}

\texttt{Voiceover: ``Covered, our dough rests and dreams of becoming sourdough gold. Patience is key.''}\\

\texttt{[scene-9]}

\texttt{[Time-lapse of dough rising in a revered spot on the countertop]}

\texttt{Voiceover: ``In a matter of hours or maybe a few more, our dough balloons like a dream.''}\\

\texttt{[scene-10]}

\texttt{[Cut to slicing the bread, revealing a perfect crumb]}

\texttt{Voiceover: ``Behold, the fruits of our four-day endeavor. A loaf with a tangy taste and airy crumb.''}\\

\texttt{[scene-11]}

\texttt{[Text on screen: ``Enjoy the Fruits of Your Labor'']}

\texttt{Voiceover: ``Indulge in this mouthwatering creation, a labor of love, from your own hands.''}\\

\texttt{[scene-12]}

\texttt{[End with an inspiring shot of the freshly baked sourdough bread, neatly sliced]}

\texttt{Voiceover: ``Now you're in the loop of home-baked magic!''}\\

\texttt{[video-script-end]}

\end{tcolorbox}

\clearpage

Lastly, we provide a sample model response collected from Olmo-2-13B model for same prompt mentioned above,

\section{Distribution of Model Responses}

\begin{table}[t]
    \centering
    \begin{adjustbox}{max width=\textwidth}
    \begin{tabular}{lccccc}
        \toprule
        & & \multicolumn{4}{c}{\textbf{Percentiles}} \\
        \textbf{Model} & 0 & 25 & 50 & 75 & 100 \\
        \midrule
        OLMo-2-13B & 1 & 86 & 130 & 243 & 553 \\
        OLMo-2-7B & 0 & 128 & 181 & 259 & 544 \\
        \midrule
        Llama-3.1-8B & 5 & 230 & 285 & 357 & 733 \\
        Llama-3.2-3B & 1 & 211 & 337 & 596 & 729 \\
        Llama-3.2-1B & 1 & 260 & 588 & 647 & 740 \\
        \midrule        
        Phi-3-med & 1 & 353 & 396 & 428 & 540 \\
        Phi-3-mini & 1 & 175 & 259 & 354 & 511 \\
        \bottomrule
    \end{tabular}
    \end{adjustbox}
    \caption{\textbf{Number of words in model response.} We report the distributions of model response length. We measure length as the number of white-space-separated words. We report the 0/25/50/75 and 100 percentile values of response length for each model considered in the analysis. All presented values in this table are calculated with the temperature value of 1.}
    \label{tab:app:len_percentile}
\end{table}

We report the distribution of the response lengths of different LLMs across different percentiles in \Cref{tab:app:len_percentile}.

\section{Variations in Model Response Lengths}
\label{sec:app:variation_in_response_len}

\begin{figure*}[t]
    \centering
    \includegraphics[width=\textwidth]{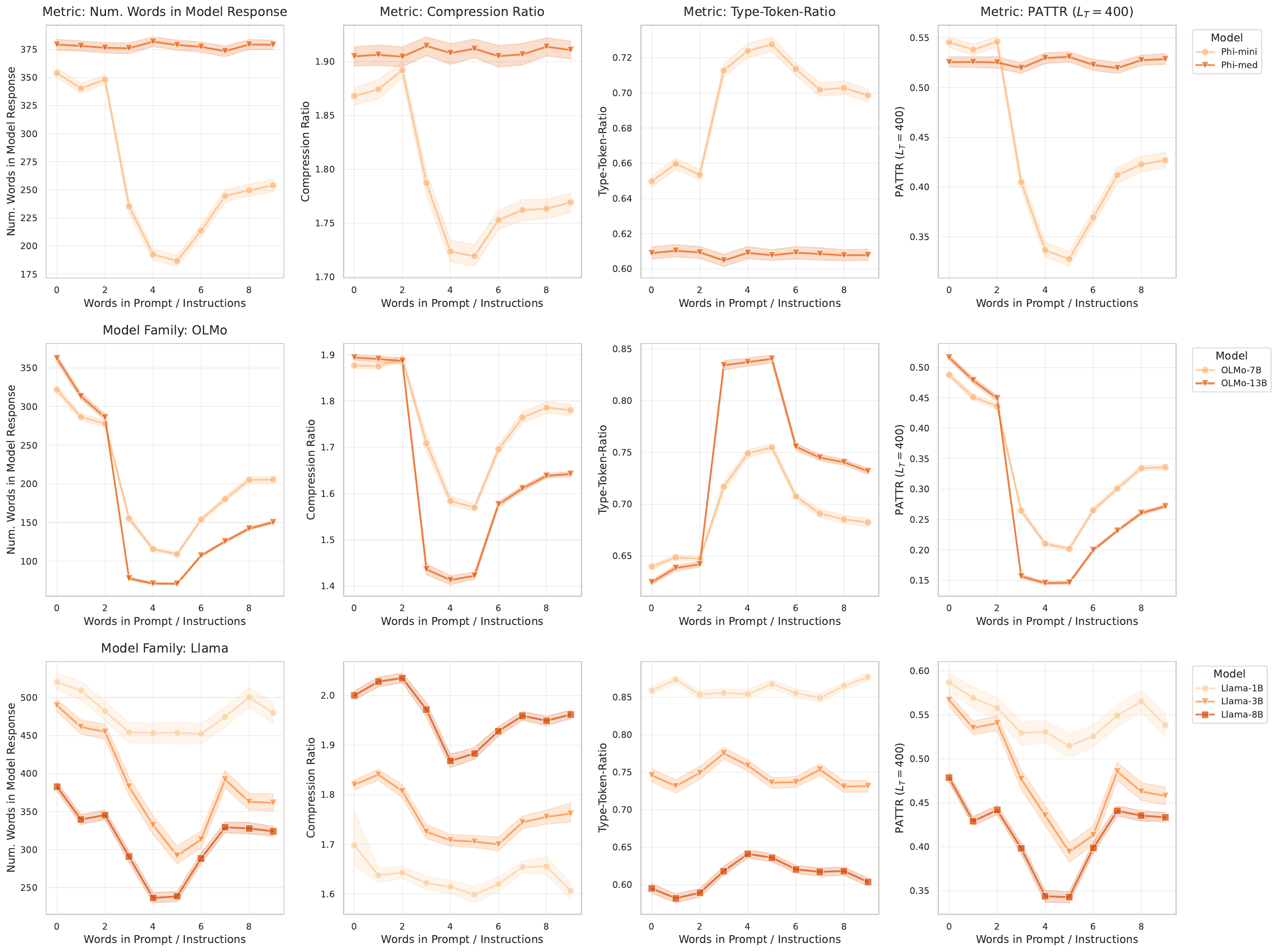}
    \caption{\textbf{Variation in response length and corresponding diversity scores.} \textit{Left to right:} response length variations, diversity scores using Compression Ratio (CR), using Type-Token Ratio (TTR), and using \methodind.}
    \label{fig:app:resp_len_vs_num_inst}
\end{figure*}

Expanding on \Cref{fig:variation_in_response_len}, we show the variations in response length and corresponding diversity scores for the Phi, OLMo, and Llama family of models in \Cref{fig:app:resp_len_vs_num_inst}.

\section{Length Bias}

\begin{table*}[t]
    \centering
    \begin{adjustbox}{max width=\textwidth}
    \begin{tabular}{lC{0.5mm}cC{0.5mm}cC{0.5mm}ccc}
        \toprule
        && \textbf{CR} && \textbf{MATTR} && \multicolumn{3}{c}{\textbf{\methodind}} \\
        \textbf{Model} && $L=128$ && $W=32$ && \textbf{$L_T=200$} & \textbf{$L_T=400$} & \textbf{$L_T=600$} \\
        \cmidrule{1-1}\cmidrule{3-3}\cmidrule{5-5}\cmidrule{7-9}
        OLMo-2-13B      && $67.33$ && $38.33$ && $28.75$ & $0.58$ & $0.08$ \\
        OLMo-2-7B       && $52.83$ && $37.58$ && $29.17$ & $0.42$ & $0.00$ \\
        \cmidrule{1-1}\cmidrule{3-3}\cmidrule{5-5}\cmidrule{7-9}
        Llama-3.1-8B    && $37.17$ && $20.17$ && $64.67$ & $0.58$ & $0.17$ \\
        Llama-3.2-3B    && $52.00$ && $7.50$ && $40.67$ & $7.75$ & $3.58$ \\
        Llama-3.2-1B    && $74.08$ && $7.17$ && $41.42$ & $28.33$ & $11.83$ \\
        \cmidrule{1-1}\cmidrule{3-3}\cmidrule{5-5}\cmidrule{7-9}
        Phi-3-med       && $43.50$ && $42.58$ && $90.00$ & $20.83$ & $1.92$ \\
        Phi-3-mini      && $48.25$ && $37.67$ && $49.75$ & $0.58$ & $0.08$ \\
        \bottomrule
    \end{tabular}
    \end{adjustbox}
    \caption{\textbf{Win rate for short responses.} We evaluate the tendency of diversity metrics to favor shorter responses. Given a pool of 10 model-generated responses for a fixed set of instructions and user prompts, we select the most diverse response using Compression Ratio (CR), Moving Average Type-Token Ratio (MA-TTR), and PATTR ($L_T \in \{200, 400, 600 \}$). The win rate represents the percentage of selected responses with a word count below the $25^{th}$ percentile of the pool. Higher win rates indicate a stronger bias toward shorter sequences. PATTR, relying on $L_T$, can achieve better robustness to length bias.}
    \label{tab:app:win_rate_full}
\end{table*}

Expanding on \Cref{tab:win_rate_small}, we show the win rates of different models from the OLMo, Llama, and Phi families in \Cref{tab:app:win_rate_full}.

\section{Data Filtration with \methodind}
\label{sec:data_filtration}

\begin{figure*}[t]
    \centering
    \includegraphics[width=\textwidth]{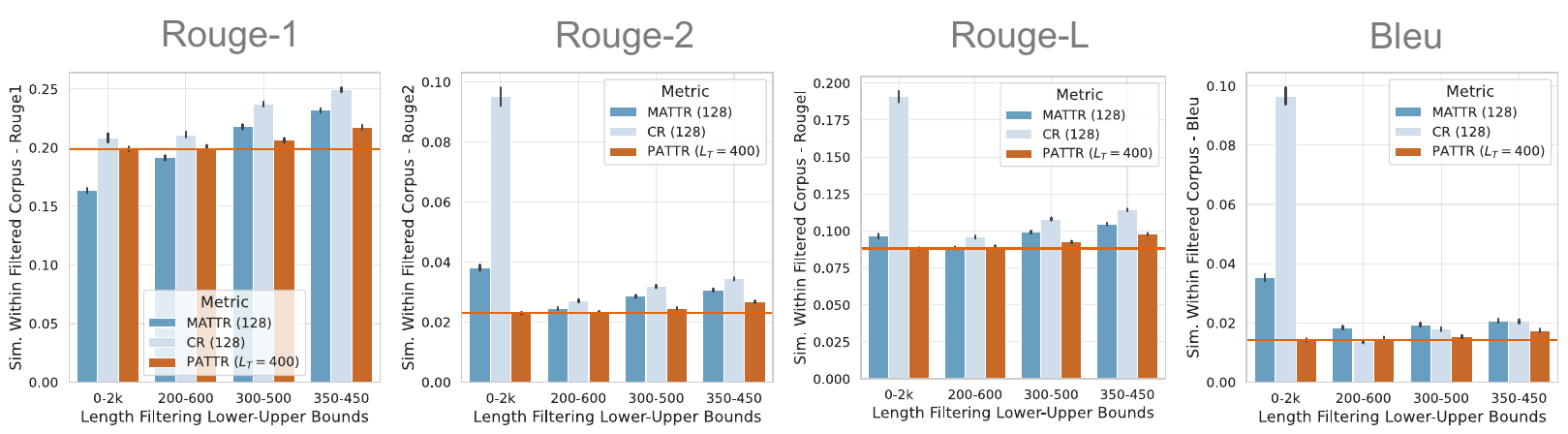}
    \includegraphics[width=\textwidth]{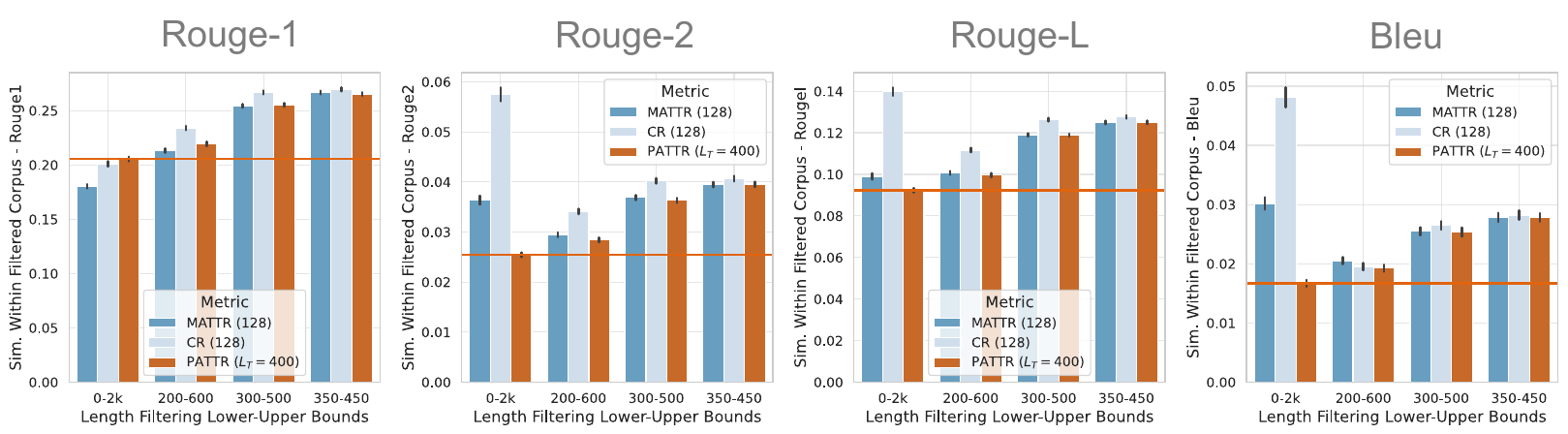}
    \caption{\textbf{Evaluation of top-100/1,000 diverse examples with pairwise similarity scores.} Average pairwise similarity scores (ROUGE-1/2/L, BLEU) for the \underline{top-100 (top row)} and \underline{1,000 (bottom row)} diverse examples selected by \methodind\ ($L_T=400$) MATTR, and CR. The $x$-axis represents different length constraints (\textit{e.g.}, 200-600: 200 $\leq$ word count $\leq$ 600). The $y$-axis shows similarity scores (lower values indicate greater diversity). The similarity scores are averaged for all seven models. The horizontal orange line represents \methodind\ without length filtering.}
    \label{fig:app:pattr_data_filter_top_100}
\end{figure*}

Supplementing \Cref{fig:pattr_data_filter_top_10}, we show the evaluations of top-100 and top-1,000 selections by \methodind, MATTR, and CR, with the target video script length set to 400 words, in \Cref{fig:app:pattr_data_filter_top_100}.

\begin{figure*}[t]
    \centering
    \includegraphics[width=\textwidth]{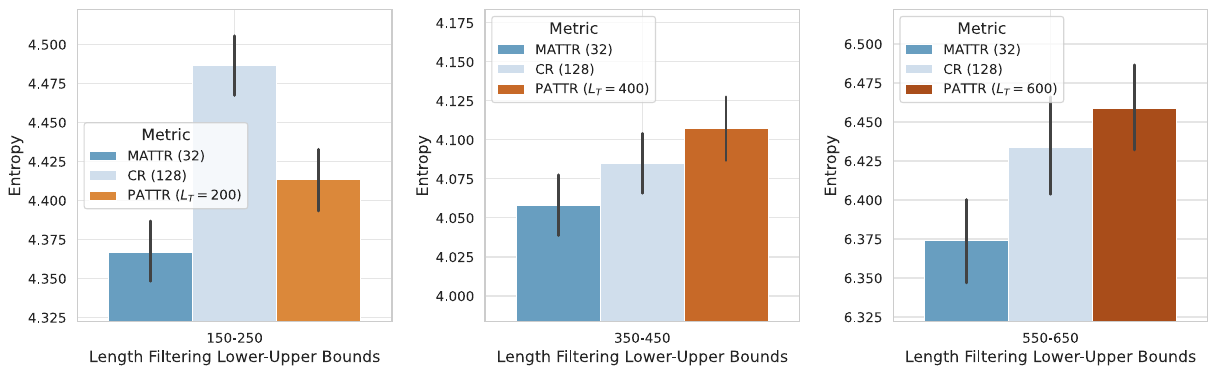}
    \includegraphics[width=\textwidth]{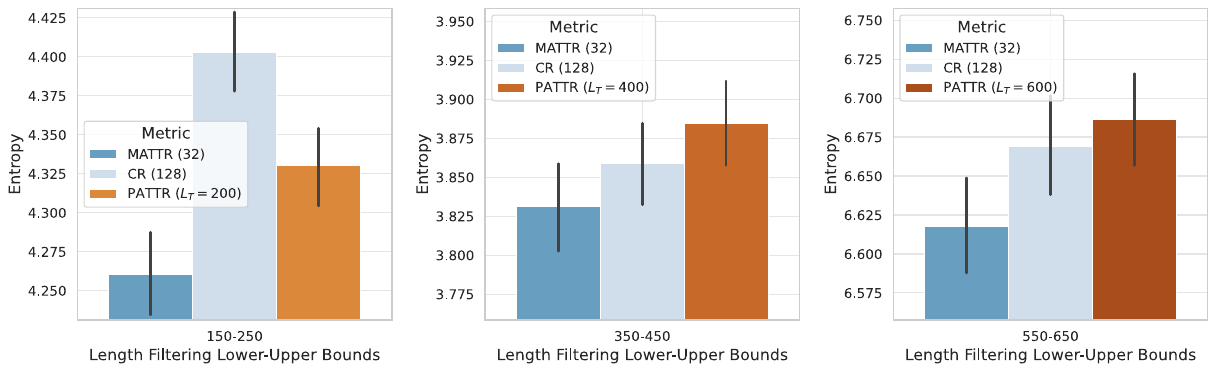}
    \caption{\textbf{Evaluation of top-1,000 diverse examples with entropy.} We measure the diversity of top-1,000 examples selected by \methodind\ ($L_T \in \{200, 400, 600\}$), MATTR (32-word window), and CR (first 128 words) with entropy based on SmolLM2-135M \underline{(top row)} and 360M \underline{(bottom row)}. Higher values of entropy represent a more diverse set of video scripts.}
    \label{fig:app:pattr_data_filter_top_1000_entropy}
\end{figure*}

Supplementing \Cref{fig:pattr_data_filter_top_1000_entropy}, we present entropy measurements using the SmolLM2-135M and SmolLM2-360M checkpoints in \Cref{fig:app:pattr_data_filter_top_1000_entropy}.

\section{Sensitivity Analysis}
\label{sec:app_sensitivity_analysis}

We provide the full version of the \Cref{tab:sensitivity_analysis_t_test} in this appendix.

\begin{table*}[t]
\centering
\small
\begin{tabular}{ccccccccc}
\toprule
\textbf{Length Filter} & \textbf{PATTR} & \textbf{MATTR} & $\Delta$ &  & \textbf{R1} & \textbf{R2} & \textbf{RL} & \textbf{B} \\
(num. words) & $L_T$ & $W$ & \textbf{Length} & \textbf{Model} & T-stat. & T-stat. & T-stat. & T-stat. \\
\midrule
0--2048 & 100 & 32 & 93.6  & Llama-3.1-8B & -1.85$^{\wedge}$ & -4.55$^{***}$ & -5.95$^{***}$ & 10.36$^{***}$ \\
0--2048 & 100 & 128 & 93.6 & Llama-3.1-8B & -1.85$^{\wedge}$ & -4.55$^{***}$ & -5.95$^{***}$ & 10.36$^{***}$ \\
0--2048 & 100 & 512 & 93.6 & Llama-3.1-8B & -1.48$^{\wedge}$ & -4.20$^{***}$ & -5.73$^{***}$ & 4.57$^{***}$ \\
\midrule
0--2048 & 275 & 32 & 275   & Llama-3.1-8B & -6.43$^{***}$ & -6.52$^{***}$ & -9.24$^{***}$ & 4.77$^{***}$ \\
0--2048 & 275 & 128 & 275  & Llama-3.1-8B & -6.43$^{***}$ & -6.52$^{***}$ & -9.24$^{***}$ & 4.77$^{***}$ \\
0--2048 & 275 & 512 & 275  & Llama-3.1-8B & -6.22$^{***}$ & -6.08$^{***}$ & -9.16$^{***}$ & 0.59$^{\wedge}$ \\
\midrule
0--2048 & 400 & 32 & 411.7 & Llama-3.1-8B & -9.27$^{***}$ & -7.52$^{***}$ & -10.96$^{***}$ & 2.02$^{*}$ \\
0--2048 & 400 & 128 & 411.7 & Llama-3.1-8B & -9.27$^{***}$ & -7.52$^{***}$ & -10.96$^{***}$ & 2.02$^{*}$ \\
0--2048 & 400 & 512 & 411.7 & Llama-3.1-8B & -9.19$^{***}$ & -7.03$^{***}$ & -10.95$^{***}$ & -0.56$^{\wedge}$ \\

\midrule
\midrule

0--2048 & 100 & 32  & 85.6  & OLMo-2-13B & -5.94 *** & -10.02 *** & -10.06 *** & -6.89 *** \\
0--2048 & 100 & 128 & 85.6  & OLMo-2-13B & -5.72 *** & -10.26 *** & -9.84 ***  & -6.61 *** \\
0--2048 & 100 & 512 & 85.6  & OLMo-2-13B & -5.72 *** & -10.26 *** & -9.84 ***  & -6.61 *** \\
\midrule
0--2048 & 275 & 32  & 260.6 & OLMo-2-13B & -5.79 *** & -12.55 *** & -12.21 *** & -11.29 *** \\
0--2048 & 275 & 128 & 260.6 & OLMo-2-13B & -5.57 *** & -12.83 *** & -11.98 *** & -10.95 *** \\
0--2048 & 275 & 512 & 260.6 & OLMo-2-13B & -5.57 *** & -12.83 *** & -11.98 *** & -10.95 *** \\
\midrule
0--2048 & 400 & 32  & 384.8 & OLMo-2-13B & -4.19 *** & -12.15 *** & -12.01 *** & -11.10 *** \\
0--2048 & 400 & 128 & 384.8 & OLMo-2-13B & -3.99 *** & -12.43 *** & -11.77 *** & -10.76 *** \\
0--2048 & 400 & 512 & 384.8 & OLMo-2-13B & -3.99 *** & -12.43 *** & -11.77 *** & -10.76 *** \\

\midrule
\midrule

0--2048 & 100  & 32  & 88.7  & Phi-3-med & 2.62 *     & 2.93 **     & -2.07 *     & -7.33 *** \\
0--2048 & 100  & 128 & 86.8  & Phi-3-med & 3.48 ***   & 4.45 ***    & -1.20$^{\wedge}$ & -6.10 *** \\
0--2048 & 100  & 512 & 87.6  & Phi-3-med & 2.42 *     & 4.45 ***    & -0.99$^{\wedge}$ & -5.14 *** \\
\midrule
0--2048 & 275  & 32  & 264.8 & Phi-3-med & 10.48 ***  & 8.74 ***    & -1.32$^{\wedge}$ & -8.31 *** \\
0--2048 & 275  & 128 & 262.9 & Phi-3-med & 12.37 ***  & 12.54 ***   & -0.28$^{\wedge}$ & -7.06 *** \\
0--2048 & 275  & 512 & 263.7 & Phi-3-med & 7.83 ***   & 12.54 ***   & -0.40$^{\wedge}$ & -5.99 *** \\
\midrule
0--2048 & 400  & 32  & 386.5 & Phi-3-med & 13.73 ***  & 12.48 ***   & -0.98$^{\wedge}$ & -8.79 *** \\
0--2048 & 400  & 128 & 384.6 & Phi-3-med & 16.08 ***  & 19.65 ***   & 0.12$^{\wedge}$  & -7.53 *** \\
0--2048 & 400  & 512 & 385.4 & Phi-3-med & 9.95 ***   & 19.65 ***   & -0.17$^{\wedge}$ & -6.41 *** \\

\bottomrule
\end{tabular}
\vspace{-1mm}
\caption{\textbf{Effect of variations in the target length.} Negative t-statistics indicate that PATTR-selected top-10 responses are more diverse. A positive $\Delta$ Length means these responses are longer than those selected by MATTR. Importantly, a positive t-stat value does not necessarily mean that PATTR is less diverse, if PATTR selected responses are longer. Rouge and Bleu metric has length bias (shorter texts can falsely appear as more diverse) and a positive t-stat. value (in the presence of positive $\Delta$Length) is likely due to the length bias of the Rouge/Bleu metric.  Significance: $^{\wedge}: p \geq 0.05$, $^{*}: p<0.05$, $^{**}: p<0.01$, $^{***}: p<0.001$. Metrics: R1 = Rouge-1, R2 = Rouge-2, RL = Rouge-L, B = Bleu.}
\label{tab:sensitivity_analysis_t_test_FULL}
\end{table*}

\end{document}